# Hybrid Supervised and Reinforcement Learning for the Design and Optimization of Nanophotonic Structures

*Christopher Yeung[1,4], Benjamin Pham[1], Zihan Zhang[2,3], Katherine T. Fountaine[4], and Aaswath P. Raman[1,5]\**

[1]Department of Materials Science and Engineering, University of California, Los Angeles, CA 90095, USA
[2]Department of Mathematics, University of California, Los Angeles, CA 90095, USA
[3]Department of Statistics, University of California, Los Angeles, CA 90095, USA
[4]NG Next, Northrop Grumman Corporation, Redondo Beach, CA 90278, USA
[5]California NanoSystems Institute, University of California, Los Angeles, CA 90095, USA
\*Corresponding Author: aaswath@ucla.edu

**ABSTRACT:** From higher computational efficiency to enabling the discovery of novel and complex structures, deep learning has emerged as a powerful framework for the design and optimization of nanophotonic circuits and components. However, both data-driven and exploration-based machine learning strategies have limitations in their effectiveness for nanophotonic inverse design. Supervised machine learning approaches require large quantities of training data to produce high-performance models and have difficulty generalizing beyond training data given the complexity of the design space. Unsupervised and reinforcement learning-based approaches on the other hand can have very lengthy training or optimization times associated with them. Here we demonstrate a hybrid supervised learning and reinforcement learning approach to the inverse design of nanophotonic structures and show this approach can reduce training data dependence, improve the generalizability of model predictions, and shorten exploratory training times by orders of magnitude. The presented strategy thus addresses a number of contemporary deep learning-based challenges, while opening the door for new design methodologies that leverage multiple classes of machine learning algorithms to produce more effective and practical solutions for photonic design.

# Introduction

Nanophotonic devices and circuits can manipulate the phase, amplitude, and polarization of light within an ultra-compact footprint. Due to these unique properties, nanophotonic devices are now widely used in a broad range of next-generation technologies and applications, such as wireless communications [1, 2], passive/active thermal management [3, 4], and optical displays for virtual or augmented reality [5, 6]. For example, to achieve arbitrary transformations to an incident wave front, metagratings or metasurfaces utilize subwavelength scattering elements and phase-shifting materials to suppress undesired diffraction orders and reroute incident power towards desired ones with high efficiency [7]. This fundamental yet versatile capability has allowed metagratings to tailor the electromagnetic spectrum and realize holograms [8, 9], beam steerers [10, 11], and ultrathin optical components [12, 13]. However, the development of new and complex nanophotonic devices, including metagratings and metasurfaces, faces tremendous bottlenecks due to the large and nonlinear design space of materials and geometries that must be explored.

To meet the increasing demand for high-performance and application-specific nanophotonic devices, a number of strategies have been proposed and deployed over the past decade to solve the inverse design problem, which is the retrieval of the optimal material and structure in response to the desired optical behavior. Conventional inverse design methods include evolutionary algorithms [14] and adjoint-based optimization [15]. Though such algorithms have been successfully applied to various photonic design problems [16], in recent years, inverse design frameworks based on artificial intelligence and deep learning have emerged and proved to be advantageous in numerous aspects. In comparison to conventional algorithms, deep learning-based methods have demonstrated superior device performance, computational efficiency, and the ability to derive new physical insights from the investigated design space [17-20]. Despite significant progress in deep learning for photonic design, many challenges and limitations still remain, in particular with each category of machine learning algorithms.

Deep and machine learning approaches can be grouped into three main classes of algorithms: supervised learning (SL), unsupervised learning (USL), and reinforcement learning (RL) [21]. SL and USL deep learning typically involves training a neural network, or related model, by using large quantities of labeled and unlabeled data, respectively. Once trained, the models can arrive at solutions orders of magnitude faster than conventional inverse design methods since the model has in principle captured a high-dimensional nonlinear function approximation between the inputs and outputs [22]. In the photonics context, SL and USL have been applied to core-shell nanoparticles [23], infrared-controlled metasurfaces [24, 25], optical thin-films [26, 27], and more. Though promising, the acquisition of large training datasets may not always be practical, particularly when a near-optimal solution may have already been found by the time sufficient training data is acquired. Additionally, when faced with optimizing for a target response outside of the training dataset, it is well-known that such neural networks struggle to identify an accurate solution [28-30]. Thus, the main limitations of both SL and USL are their training data requirements and poor model generalizability, while their core strengths are post-training computational speed and efficiency.

On the other hand, in RL, a model known as an "agent" is trained through trial-and-error by evaluating its actions within an environment and producing a corresponding reward (*e.g.*, a favorable action will be met with a positive reward and vice-versa). Therefore, there is no need for any training data, and RL is not bound to "prior knowledge" (as in SL and USL) since the method is inherently capable of self-exploration and exploitation. As a result, several works have demonstrated RL for the design of photonic devices and components [31-34]. Recent results show that RL can achieve lower variance and higher performance than the adjoint method as well as adjoint optimization-enhanced generative networks that rely on little to no training data [35, 36]. Another unique property of RL (compared to conventional optimization algorithms) is the ability to reuse and apply a solution (or trained model) to other similar problems via transfer learning [36, 37]. However, due to the exploratory nature of RL, it is substantially slower than SL or USL models that are trained on readily-available datasets, with problems taking up to days [37, 38] or even an entire month [39-41] to solve.

In this work, we aim to combine the strengths of supervised and reinforcement learning for the design and optimization of nanophotonic structures, which in turn addresses their individual weaknesses to achieve superior design performance. By utilizing a relatively small training dataset (~5,000 data points), we first train a convolutional neural network (CNN) and leverage its high speed to inverse design a silicon-on-insulator metagrating in response to an arbitrary target electric field profile (shown in Figure 1a). After generating the metagrating design, we refine device performance further than that achievable through the SL model alone by passing the design through an RL process (shown in Figure 1b). The RL agent sequentially adjusts the metagrating design by either making grating widths smaller or larger to minimize the difference between the target and output electric field patterns. Through this workflow, we extend the generalizability of the SL model's predictions, while simultaneously reducing RL training time by providing the algorithm with a better starting point for training. We validate our results by comparing the individual and combined application of SL and RL for a variety of design targets.

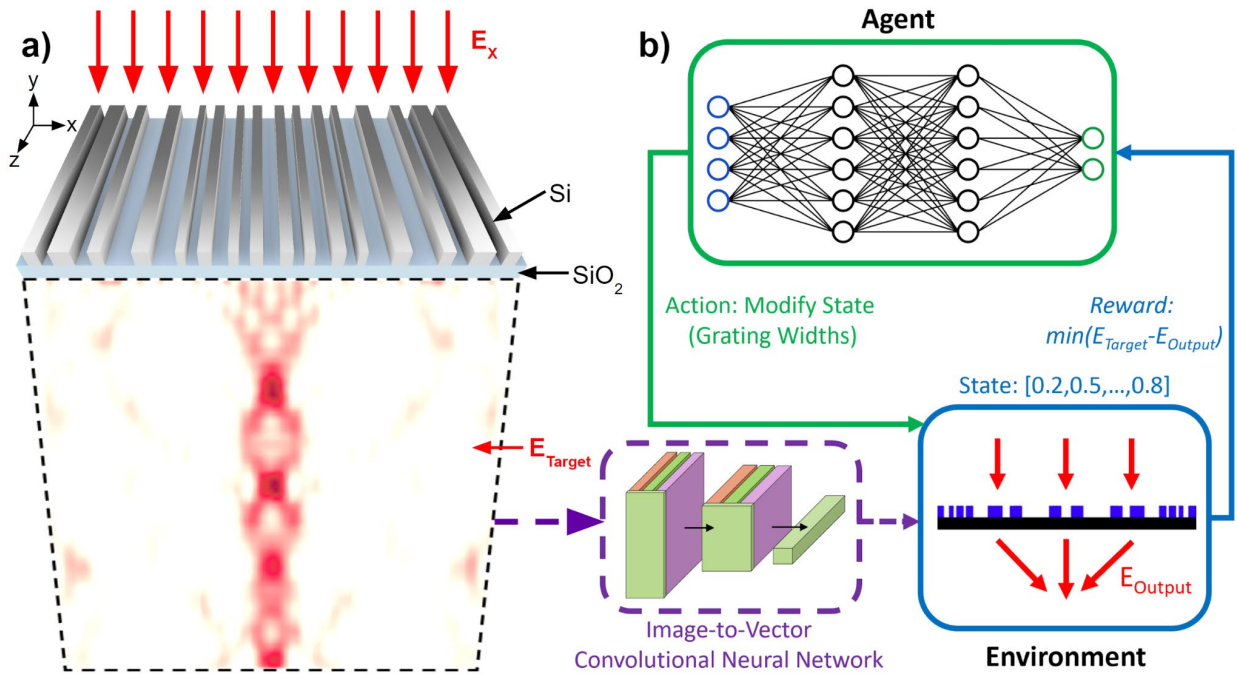

**Figure 1:** Inverse design of transmissive metagratings. (a) Illustration of design problem. Incident light (1.5 μm) transmitted through an unknown metagrating design, consisting of patterned strips of Si on $SiO_2$, produces a particular electric field profile ($E_{Target}$). (b) Deep learning design framework integrating supervised and reinforcement learning. A convolutional neural network predicts an approximate metagrating design, then reinforcement learning improves metagrating performance by modifying design parameters until $E_{Target}$-$E_{Output}$ is minimized and the unknown design is found. This joint method allows us to overcome the limitations associated with each individual deep learning algorithm.

# Results and discussion

**Supervised learning data preparation and model evaluation**

We first evaluated the ability of a supervised learning model to perform the inverse design of metagratings by building a neural network model that was trained on a relatively small training dataset. As shown in Figure 1b, our design problem is graphically illustrated as a 2D metagrating captured along the XY-plane of the device, where a plane wave (1.5 μm under X-polarization) was injected at normal incidence above the structure (propagating along the Y-axis), and a transmissive electric field profile was generated below the structure. The metagrating consisted of patterned strips of Si on $SiO_2$ (of uniform heights/thicknesses) that extended infinitely along the Z-axis and maintained unit-cell periodicity along the X-axis. A unique metagrating design was represented by 13 Si strips with individual widths that varied from 0 to 800 nm in 200 nm increments (with 0 nm indicating the absence of the strip). Thus, the total number of possible parameter combinations in the design space was on the order of $10^9$. In addition, symmetry boundary conditions were applied to the center of the structure along the Y-axis. Using this setup, 5,000 sets of metagrating parameters were randomly generated and simulated in MaxwellFDFD [42] to capture their electromagnetic responses for deep learning. Here, we calculated the electric field magnitudes ($E = \sqrt{E_x^2 + E_y^2 + E_z^2}$) below the metagrating structure, and utilized this profile as image inputs for the deep learning model. We note that this method of representing the target E-field in principle allows a more complex specification of phase and amplitude, and enables a designer to easily "draw" the required device response for a given application.

Next we trained a CNN in the TensorFlow framework by using the E-field images as inputs and grating widths as outputs. During hyperparameter tuning and model training, we note that two types of loss behaviors were observed. As shown in Figure 2a, the model frequently overfitted while the training and validation losses diverged. When additional measures were taken to reduce overfitting (*e.g.*, dropout, neural architecture search, and data augmentation or noise reduction), model accuracy was compromised and underfitting occurred (shown in Figure 2b). This suggests that regardless of model complexity, there were insufficient samples available to train the CNN [43]. To further assess the limitations of SL with insufficient training data, we tested the model's ability to inverse design metagratings with target E-field profiles that were associated with known designs. Figure 2c presents several example design targets: a single focal point (top), collimated beam (middle), and dual collimated beams (bottom), all of which were withheld from the training dataset. Each target was passed into the CNN that had the least overfitting and lowest validation loss (from Figure 2b; detailed in the Supplementary Material), then the model's output designs were simulated for final verification. Simulated designs are shown in Figure 2d, where we observe only marginal similarities between the target and designed E-field profiles.

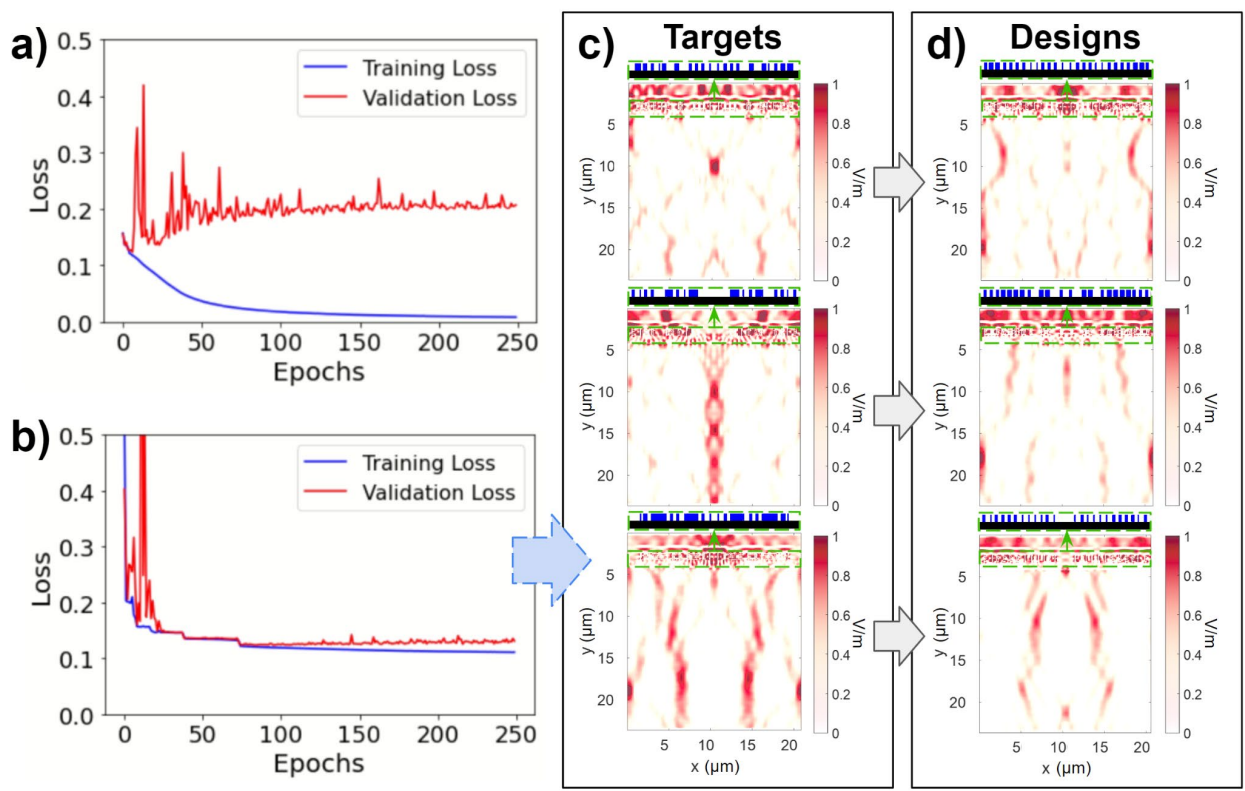

**Figure 2:** Supervised learning (SL) for the inverse design of metagratings. Inputs to the neural network were E-field images and outputs were device parameters. (a,b) Indications of training a model with insufficient data. Training and validation losses either (a) overfit or (b) underfit regardless of model complexity. Using the best achievable model, (c) example design targets were tested, which consist of metagratings with various E-field profiles: a single focal point (top), collimated beam (middle), and dual collimated beams (bottom). Targets were known designs that were withheld from the training dataset. (d) Designs predicted by a model trained with insufficient training data. Inset images show the corresponding metagrating designs.

**Reinforcement learning configuration and evaluation**

After identifying cases where the SL model failed to generalize beyond the training data, we investigated the effectiveness of RL towards performing the same design task outlined in the previous section. Here, we implemented the proximal policy optimization (PPO) algorithm [44], which previously demonstrated world championship winning results in the area of electronic sports [45]. To apply PPO to our particular design problem, the following components and configurations were required: action definitions, interactions between states and environment, and the reward function. Two actions were defined for each optimizable element (to either increase or decrease the grating width by 200 nm), yielding a total of 26 actions. At the initial timestep ($t$), the state of the metagrating was represented as the following vector: $s_t = [s_1, s_2, ..., s_{13}]$. After performing an action in the next timestep, the state was modified accordingly. For instance, given an action $a_1$ (up to $a_{26}$), which modifies the first position within the state, the following interaction takes place: $s_{t+1}(a_1) = [s_1+0.2, s_2, ..., s_{13}]$. State values were normalized from 0 to 1 for ease of training. At each timestep, the state was passed into the electromagnetic solver (MaxwellFDFD), which simulated the structure to produce an output image. We then compared the output image to the user-specified target image by using the structural similarity index (SSIM) to quantify the difference between the two images (*i.e.*, an SSIM of 0 indicates a perfect match between target and output).

In order for the agent to learn the best actions to take under specific states, a reward function was required, which penalizes or rewards the agent based on its action/state responses. To calculate the reward function, we stored the SSIM values of the previous timestep during training, such that the next action was rewarded positively if ΔSSIM was negative and vice-versa. In other words, we incentivised the agent to perform an action that produces an output image with higher similarity to the target image. In Tables S1-S3 of the Supplementary Material, we performed hyperparameter tuning on the PPO algorithm, and evaluated numerous linear and nonlinear reward functions. Here, we observe that the nonlinear reward functions (*e.g.*, sigmoidal) performed significantly better than linear functions, particularly during the early stages of training, where we exponentially rewarded/penalized actions in response to larger changes in ΔSSIM. To expedite our RL trials, we note that the reward function and hyperparameter tuning were performed using a custom-made arbitrary function regression environment (detailed in the Supplementary Material) with the same amount of parameters in our design space, rather than the MaxwellFDFD-integrated environment.

After optimizing our reward function, we trained the RL algorithm using the same targets shown in Figure 2 for 5,000 episodes and 20 timesteps per episode (which is greater than the amount of timesteps needed to reach the known designs). As the starting point for training, an initial state of all 200 nm wide gratings was used for each target (*i.e.*, $s_t = [0.2_1, 0.2_2,...,0.2_{13}]$). To produce a larger sample size, three designs were generated for each target, and each design took approximately one week of training. Calculations were run on a machine with a 16-core 3.40 GHz processor and 64 GB of RAM. Training results are presented in Figure 3, where we generally observe poor convergence and design performance. Since model training is initialized using random weights and actions, we note that each RL run produced a different design after 5,000 episodes. The results here indicate that longer training times or additional model optimization were required.

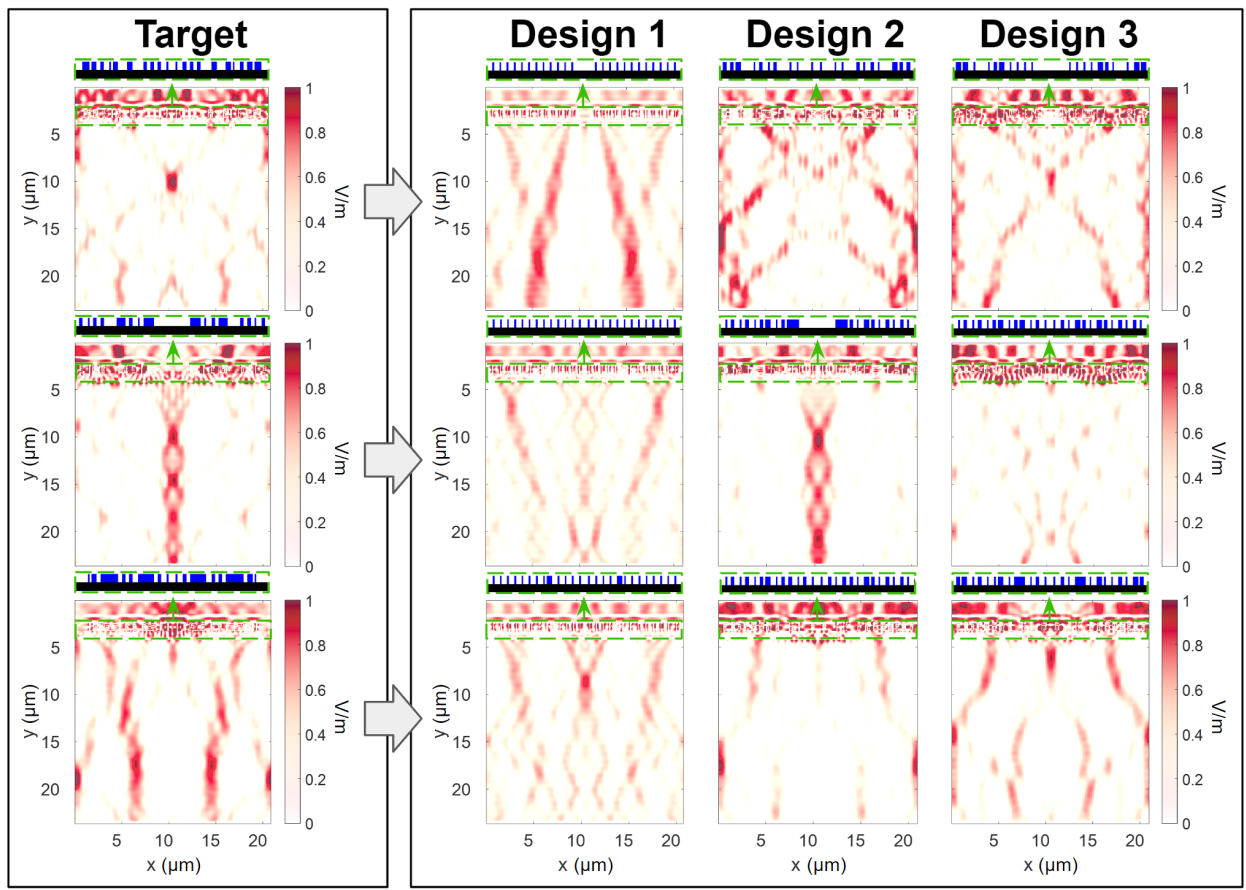

**Figure 3:** Reinforcement learning (RL) for the inverse design of metagratings. The example design targets were fed into the proximal policy optimization (PPO) algorithm and trained for 5,000 episodes (~1 week) using a fixed metagrating starting design. Three designs were generated for each target and poor design convergence can be observed; indicating longer training times or additional model optimization were required. Inset images show the corresponding metagrating designs.

**Combining supervised and reinforcement learning**

In this section, we combined SL and RL in an effort to simultaneously address the previously demonstrated limitations of each deep learning method. In particular, we used the CNN-predicted metagrating designs from Figure 2 as starting points for RL, and presented the results in Figure 4. In Figure 4, we observe that after 5,000 episodes, the SL+RL approach produced designs with significantly higher accuracy and performance than the RL-only results (in relation to the input targets). From a qualitative standpoint, the SL+RL designs exhibited the sought behaviors, including: a single focal point (top), collimated beam (middle), and dual collimated beams (bottom). The results here indicate that the SL model was able to assist the RL algorithm by reducing the time required for the latter to obtain an adequate solution. However, we note that the dual collimated beam designs only produced faint signals or low-power E-fields, which suggests that the degree of accuracy may depend on the complexity of target design, since such targets may possess fewer solutions (or require more precise solutions) in comparison to simpler E-field profiles. To this end, longer training times may produce results with greater accuracy, which is beyond the scope of the current study. Furthermore, we note that the number of episodes used in this study was selected based on predetermined minimum requirements necessary to produce meaningful improvements to the CNN-predicted designs. We therefore anticipate that a CNN model trained on a larger dataset may reduce the number of RL episodes required for the optimization to converge, which may also be explored in future works.

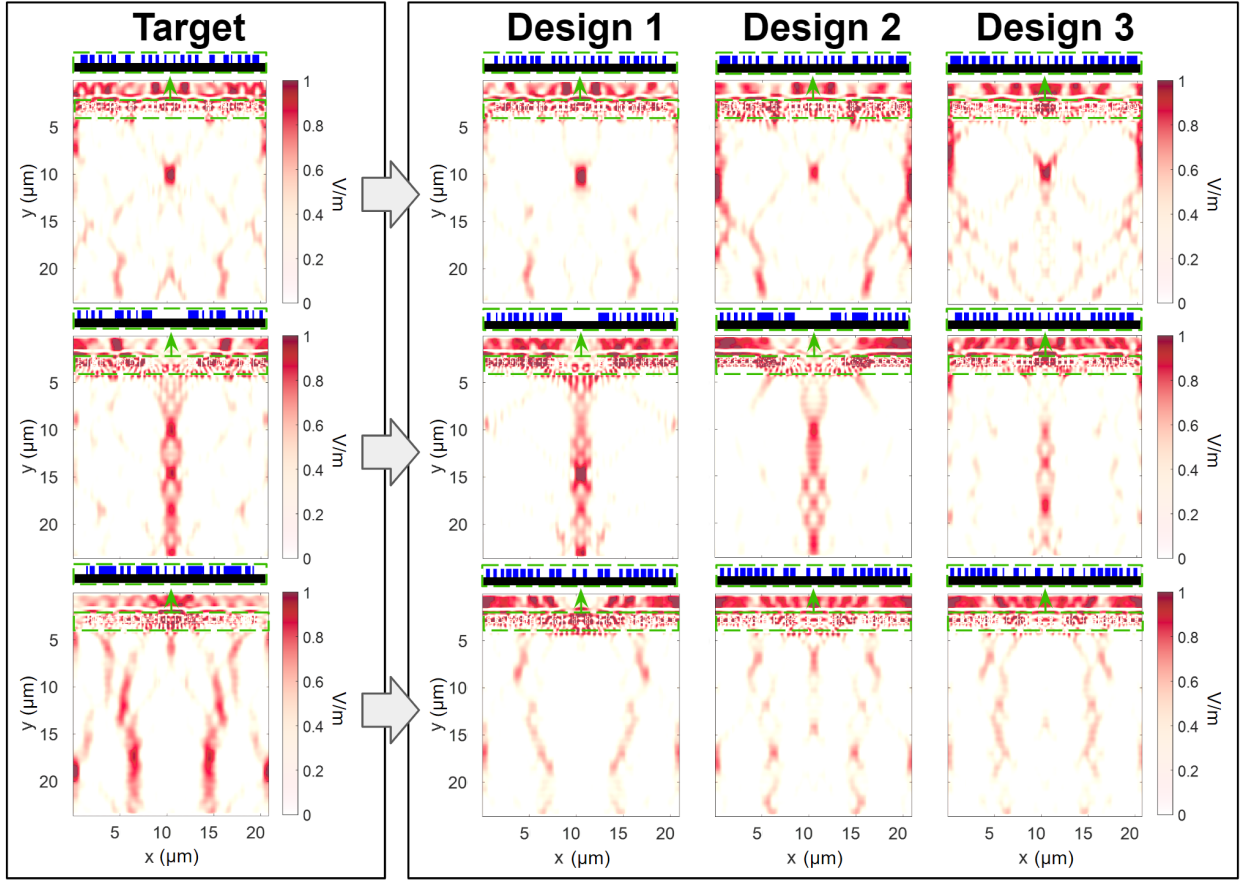

**Figure 4:** Combining SL and RL for the inverse design of metagratings. Pre-trained CNN predictions were used as the starting point for RL, which subsequently ran for 5,000 episodes. In comparison to SL- and RL-only predictions, the joint approach produced superior accuracy and converged in less time. The degree of accuracy appears to depend on the complexity of the input target. Inset images show the corresponding metagrating designs.

In Figures 5a-c, we compared the SSIM values of the SL-only, RL-only, and SL+RL designs, respectively. Figure 5a presents the SSIM of the design outputs from the best three CNN models found during model optimization. Figures 5b and 5c show the SSIM of the RL designs corresponding to Figure 3 and Figure 4, respectively. We observe that the SL+RL designs resulted in 4 times less variance (standard deviation < 1%) and 75% higher performance (lower SSIM) than SL-only designs. The SL+RL designs also had 10 times less variance and over 30% higher performance than the RL-only designs in the same number of episodes. Therefore, in this study, we show that RL can be used to generalize the predictions of an SL model by exploring beyond the latter's training dataset, while SL can assist RL in its search process by using a relatively small training dataset to capture high-level features first. We note that prior attempts at combining SL and RL required iterating through pre-trained or weight-fixed model predictions [46], which does not fit the contemporary definition of RL, where the model learns (and optimizes its weights) from the trial-and-error process itself. Thus, to our knowledge, this is the first study to explore the simultaneous application of SL and RL for nanophotonic design. Additionally, since our goal was to demonstrate the advantages of merging different deep learning techniques, we note that our implementation of RL can be enhanced with many future improvements to reduce training times further. For example, more advanced action definitions can be applied to simultaneously adjust multiple elements rather than a single element at a time, further reward function engineering may yield faster convergence rates, and parallelizing the environment or electromagnetic solver can also expedite state evaluations tremendously. Importantly, with our combined SL and RL approach, we show that training data requirements can be reduced, and model predictions do not have to be limited by the initial dataset, all by strategically harnessing the strengths of distinct deep learning algorithms and methodologies.

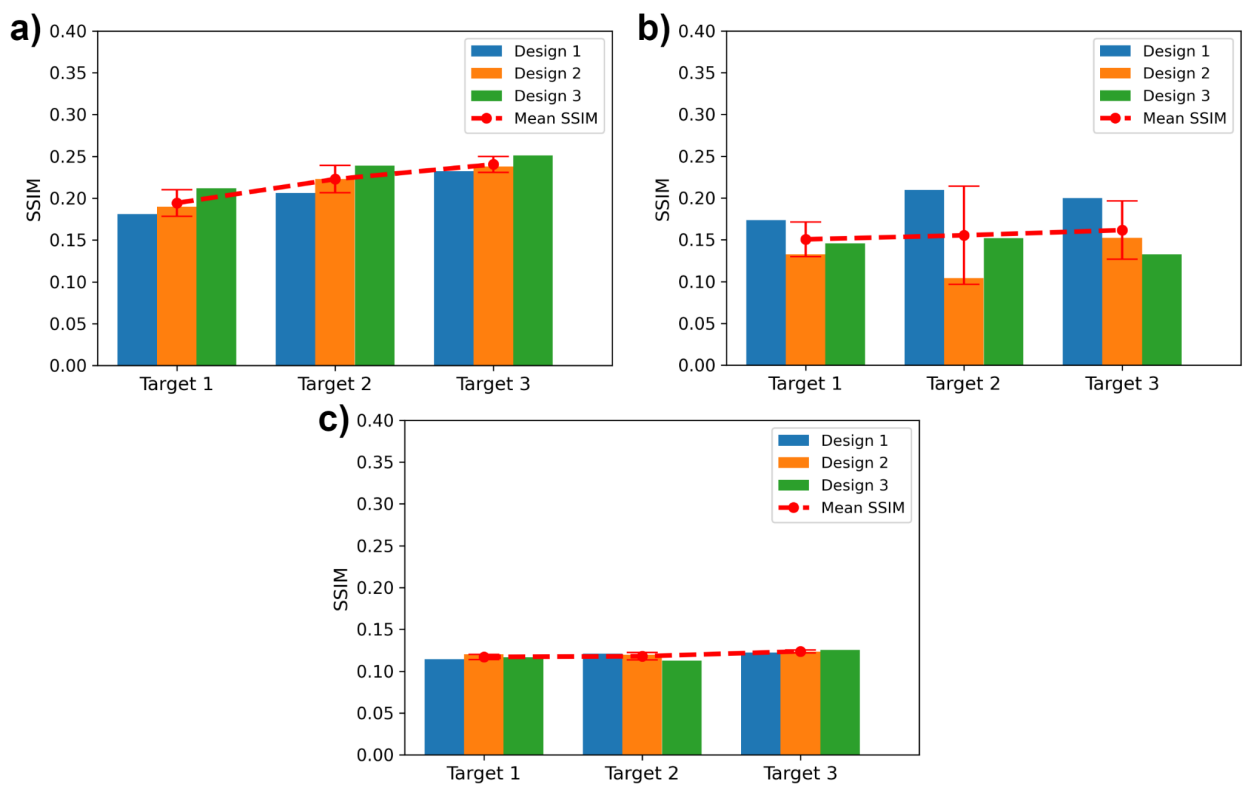

**Figure 5:** Inverse design performance comparison. Structural similarity index (SSIM) values of the generated designs were used to evaluate the performance of the (a) SL-only, (b) RL-only, (c)

and SL+RL models. Higher performance is indicated by a lower SSIM. In comparison to SL- and RL-only models, SL+RL predictions produced significantly lower variance and mean SSIM.

## Conclusions

In summary, we present a multi-class deep learning strategy for nanophotonic circuit and device inverse design that combines distinct machine learning algorithm classes, particularly supervised learning (SL) and reinforcement learning (RL), to extend the capabilities of each individual method. Notably, SL suffers from its large training data requirements and inability to generalize well beyond its training dataset, which we highlight here in the context of metagrating design. On the other hand, RL provides explorative design possibilities without the need for training data, but at the cost of very lengthy training times (days to months). To overcome their respective limitations, we trained an SL model with a relatively small dataset to identify a high-level solution, then applied RL to achieve a more precise solution. Our results show that SL+RL designs (of metagratings with arbitrarily-defined E-field profiles) resulted in 4 times less variance (standard deviation < 1%) and 75% higher performance than SL-only designs. In comparison to RL-only designs, SL+RL achieved over 10 times less variance and over 30% higher performance in the same number of episodes or training cycles. Our work therefore highlights the potential of combining multiple classes of machine learning to produce more effective and practical solutions for photonic design. We firmly believe these results will inspire the development of further hybrid algorithms or approaches that address major contemporary challenges associated with the applications of deep learning for photonic device and circuit optimization.

## Supplementary Material

Included in the supplementary material are details regarding the supervised learning model and reinforcement learning optimization.

# Acknowledgements

This work used computational and storage services associated with the Hoffman2 Shared Cluster provided by UCLA Institute for Digital Research and Education's Research Technology Group.

# Funding Sources

This work was supported by the National Science Foundation (NSF CAREER) under Grant No. 2146577, the Sloan Research Fellowship from the Alfred P. Sloan Foundation and the DARPA Young Faculty Award (#W911NF2110345).

# Conflicts of interest statement

The authors declare no conflicts of interest.

# Supplementary Material

**Convolutional Neural Network (CNN) Model and Training**

In this work, we trained a supervised deep learning model (CNN), on a dataset of 5,000 image-label pairs, where the images were E-field profiles and the labels were metagrating design parameters. The final optimized model, which we obtain using neural architecture search (AutoKeras), consisted of four convolutional layers with batch normalization, leaky ReLU activation, average pooling, L2 regularization, and dropout with 20% probability. Convolutional layers began with 64 filters (3×3 window size) and were doubled at each subsequent layer. Input images were 270×270 pixels resized to 64×64, and output parameters were 1×13 dimension vectors. Convolutional layers were followed by two fully-connected or dense layers with 100 and 13 neurons, respectively. Training and validation losses (shown in Figure 2b of the main text) were $1.1\times10^{-1}$ and $1.3\times10^{-1}$, respectively. Converging but relatively large losses present evidence of underfitting and insufficient training data.

**Proximal Policy Optimization (PPO) Training**

We tested and optimized reinforcement learning (RL) training parameters for the inverse design of metagratings with arbitrarily-defined E-field profiles. To maximize training time, since the application of the MaxwellFDFD-integrated environment can potentially take days to converge, we performed hyperparameter tuning and model validation using a custom-made arbitrary function regression environment (which took ~1 hour of training time). Within the environment, we simply define a starting state and corresponding actions for each state index that either increase or decrease the state value. Actions are rewarded based on the difference between the new state and an arbitrarily defined target function, which we quantified using ΔMSE (or mean-squared error). Thus, similar to the use case presented in the main text (where ΔSSIM was used to measure image similarity), we used the gradient of the merit function to reward or penalize the agent.

In Table S1, we present the tuned PPO hyperparameters, evolution of average rewards vs. episodes, and the final test result at the last episode which compares the target function (blue) against the RL-optimized solution (orange). For hyperparameter tuning, we began with a model of predetermined parameters as the control (Model 1) and modified parameters individually (highlighted in yellow) to evaluate their effects on model performance and convergence rate. We observe that the control model already produces an output that is relatively close to the target function. As expected, decreasing the number of training episodes and increasing the number of timesteps decreases and increases model accuracy, respectively, thus validating our RL implementation. Though in principle these parameters can be increased indefinitely, we note that in order to balance model performance and training time, we sought to minimize both the number of episodes and timesteps in the final implementation. Next, we determined the optimal conditions for K epochs, learning rate, and policy update frequency in terms of timesteps, which were 3, 0.004, and 750, respectively.

Table S1. Hyperparameter tuning for proximal policy optimization.

| Model | Episodes | Timesteps | Update Timestep | Learning Rate | K Epochs | Reward Plot | Test Result |
|---|---|---|---|---|---|---|---|
| 1 | 1000 | 1000 | 500 | 0.004 | 4 | 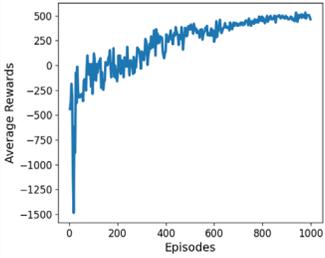 | 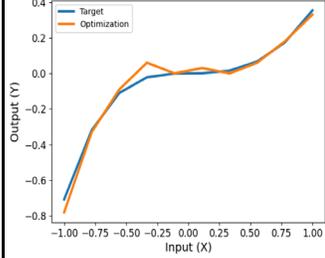 |
| 2 | 100 | 1000 | 500 | 0.004 | 4 | 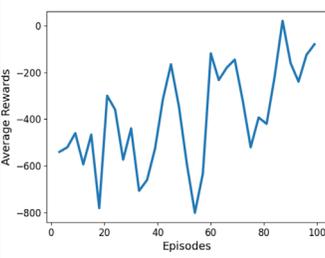 | 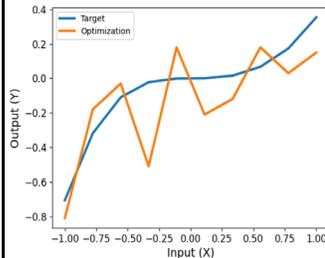 |
| 3 | 1000 | 3000 | 1500 | 0.004 | 4 | 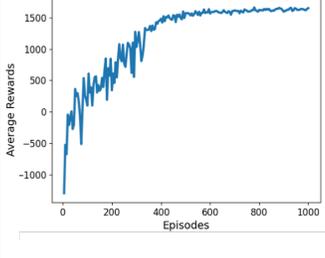 | 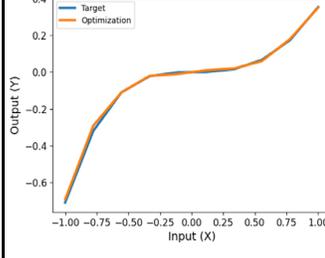 |
| 4 | 1000 | 1000 | 500 | 0.002 | 4 | 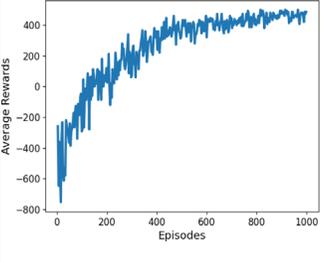 | 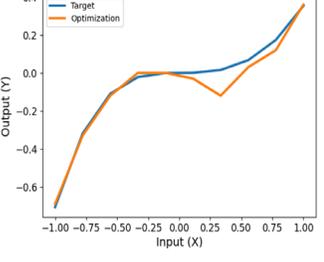 |
| 5 | 1000 | 1000 | 500 | 0.006 | 4 | 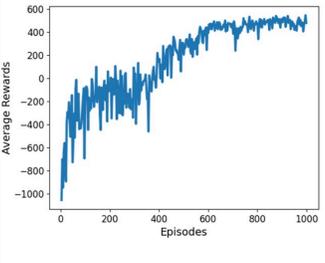 | 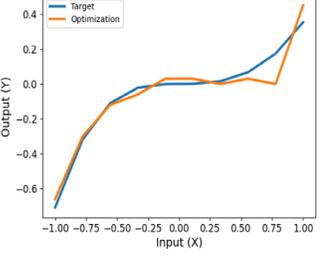 |
| 6 | 1000 | 1000 | 500 | 0.008 | 4 | 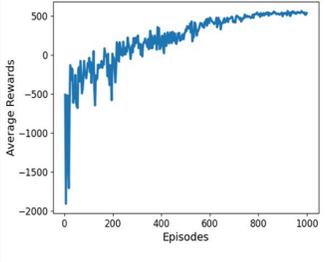 | 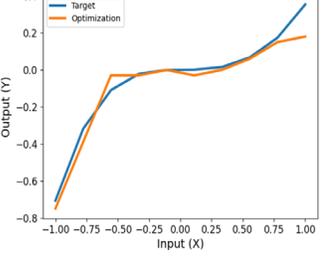 |
| 7 | 1000 | 1000 | 250 | 0.004 | 4 | 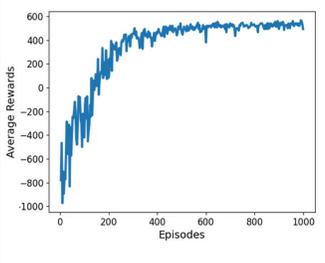 | 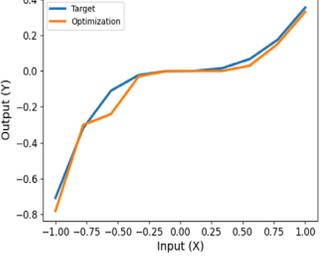 |
| 8 | 1000 | 1000 | 750 | 0.004 | 4 | 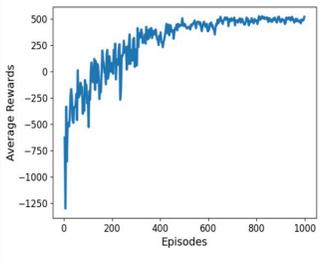 | 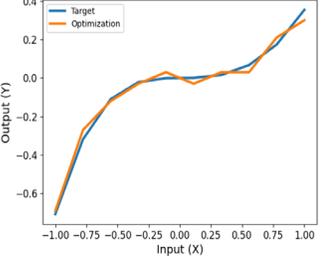 |
| 9 | 1000 | 1000 | 500 | 0.004 | 5 | 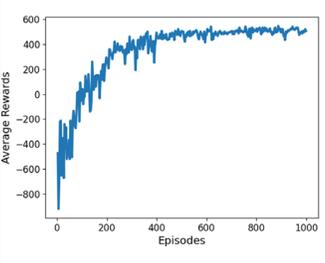 | 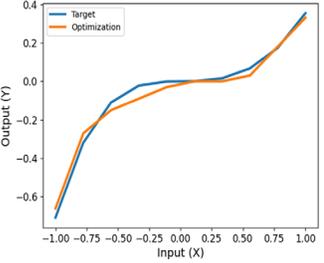 |

| | | | | | | Reward Plot | Test Result |
|---|---|---|---|---|---|---|---|
| 10 | 1000 | 1000 | 500 | 0.004 | 3 | 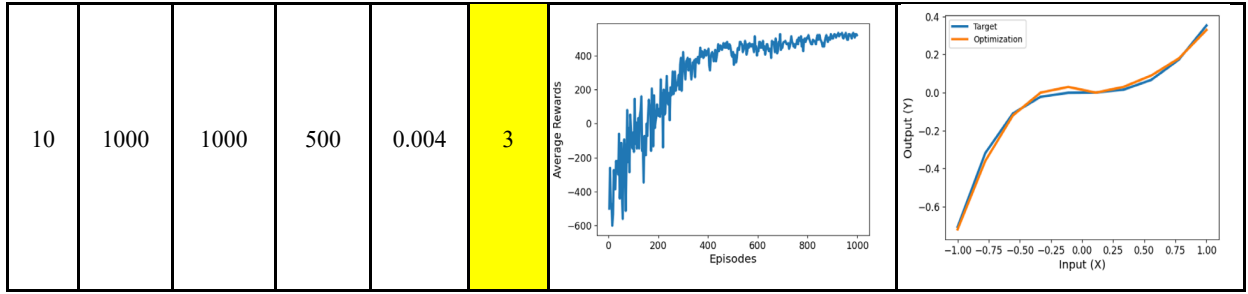 | |

After tuning the algorithm's parameters, we investigated the results of various reward functions (Table S2). Here, we experimented with various linear and nonlinear functions with the overarching goal of having positive rewards if ΔMSE < 0 and negative rewards if ΔMSE > 0, which indicated that the solution was converging or diverging, respectively. For this set of tests, the optimized parameters from the previous step were used, and maximum episodes and timesteps were both set to 1000. First, we applied a basic step function that uniformly assigned a reward of +10 if ΔMSE < 0 and -10 if ΔMSE > 0 (Model 1) and found the reward to exhibit much instability during training. We note that this is the same reward function used in Table S1. Subsequent testing revealed that reward functions with continuous and convex transitions between positive and negative rewards yielded the fastest convergence rate and highest accuracy. The final reward function was therefore a variant of the sigmoid function that is appropriately scaled to our merit function gradient.

**Table S2.** Reward function testing and optimization.

| Model | Reward Function (R) | Final MSE | Reward vs ΔMSE | Reward Plot | Test Result |
|---|---|---|---|---|---|
| 1 | **Step Function** $\Delta MSE < 0$: $R = 10$ $\Delta MSE > 0$: $R = -10$ | 0.00469161 | | | |
20

| | | | Reward vs MSE Delta | Average Rewards vs Episodes | Output vs Input |
|---|---|---|---|---|---|
| 2 | **Sigmoid 1** $R = 0.5 - \dfrac{1}{1 + e^{-1000*\Delta M}}$ | 0.010 07895 | 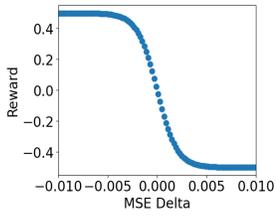 | 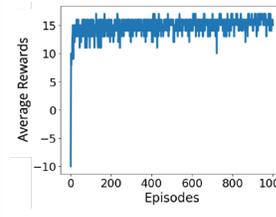 | 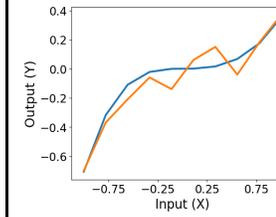 |
| 3 | **Sigmoid 2** $R = 0.5 - \dfrac{1}{1 + e^{-500*\Delta M}}$ | 0.002 73392 | 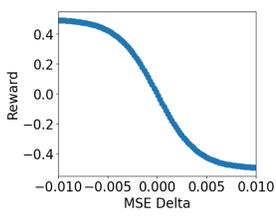 | 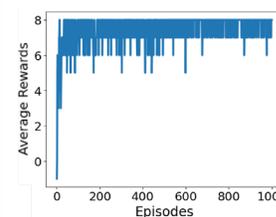 | 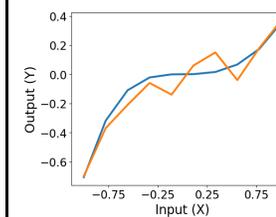 |
| 4 | **Sigmoid 3** $R = 0.5 - \dfrac{1}{1 + e^{-450*\Delta M}}$ | 0.010 85479 | 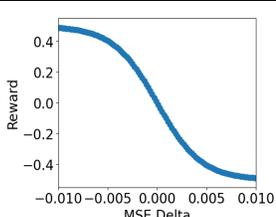 | 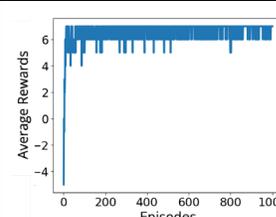 | 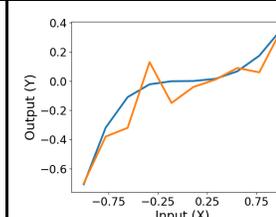 |
| 5 | **Sigmoid 4** $R = 0.5 - \dfrac{1}{1 + e^{-400*\Delta M}}$ | 0.002 61707 | 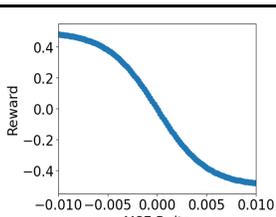 | 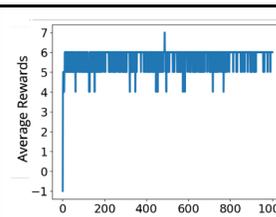 | 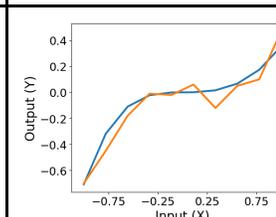 |
| 6 | **Sigmoid 5** $R = 0.5 - \dfrac{1}{1 + e^{-1200*\Delta M}}$ | 0.006 35277 | 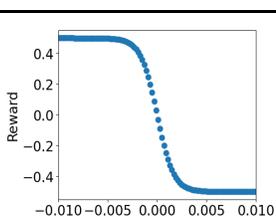 | 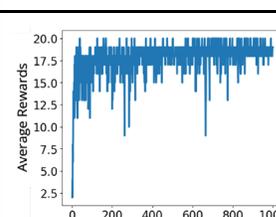 | 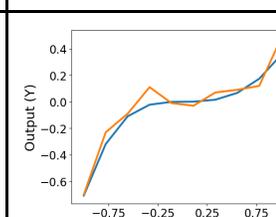 |
| 7 | **Quadratic** $\Delta MSE < 0$: $R = 10^6 * \Delta MSE^2$ $\Delta MSE > 0$: $R = -10^6 * \Delta MSE^2$ | 0.012 48142 | 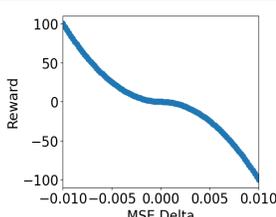 | 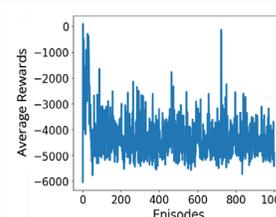 | 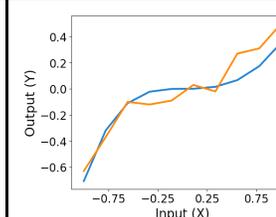 |

| # | Reward Function | Final MSE | Reward vs MSE Delta | Average Rewards vs Episodes | Output vs Input |
|---|---|---|---|---|---|
| 8 | **Linear**<br>$\Delta MSE < 0$:<br>$R = 10 - 10^6 * \Delta MSE$<br><br>$\Delta MSE > 0$:<br>$R = -10 - 10^6 * \Delta MSE$ | 0.00136640 | | | |
| 9 | **5th Power**<br>$\Delta MSE < 0$:<br>$R = 10 - 10^6 * \Delta MSE^5$<br><br>$\Delta MSE > 0$:<br>$R = -10 - 10^6 * \Delta MSE^5$ | 0.00274034 | | | |
| 10 | **Tangent**<br>$R = -tan(100 * \Delta MSE)$ | 0.00669858 | | | |
| 11 | **Step+Sigmoid**<br>$\Delta MSE < 0$:<br>$R = 10 - \dfrac{e^{\Delta MSE}}{1 + e^{\Delta MSE}}$<br><br>$\Delta MSE > 0$:<br>$R = -10 - \dfrac{e^{\Delta MSE}}{1 + e^{\Delta MSE}}$ | 0.00605135 | | | |
| 12 | **Final (Sigmoid)**<br>$R = \dfrac{1000}{1 + e^{20 * \Delta MSE}} - 500$ | 0.00109759 | | | |

Lastly, we tested our optimized model parameters and reward function on various targets in order to validate the agent's ability to generate actions that converged towards the solutions of a wide range of problems. As seen in Table S3, we tasked the agent with target functions of various shapes, and the agent was able to successfully optimize the states in order to match each target. We note that some functions were less accurate than others (*e.g.*, Target 6) and would require longer training times to produce a near-perfect match, since the degree of accuracy is determined by the starting point of RL training as well as the complexity of the target.

**Table S3.** Testing the optimized reinforcement learning model on various target functions.

| Test | Reward Plot | Test Result |
|---|---|---|
| 1 | 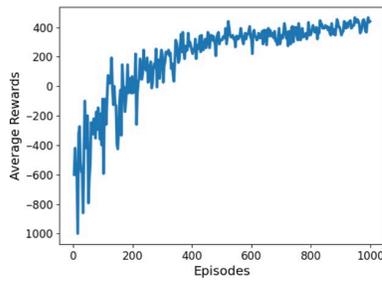 | 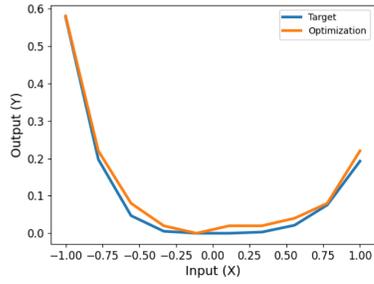 |
| 2 | 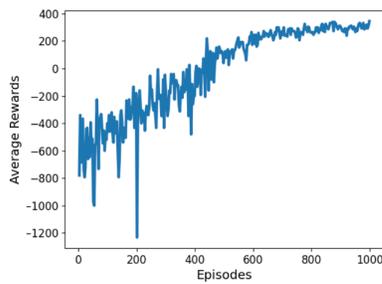 | 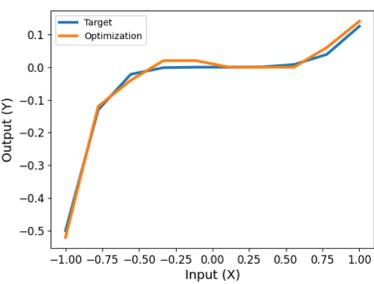 |
| 3 | 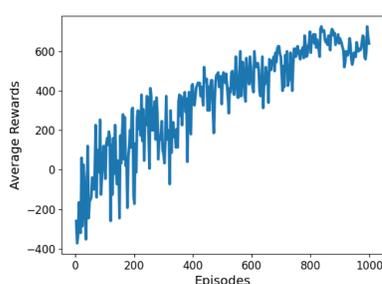 | 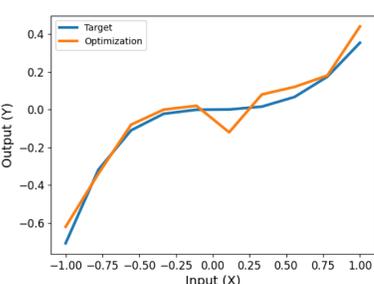 |
| 4 | 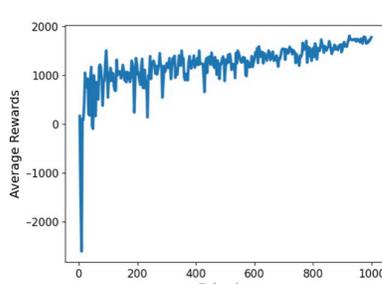 | 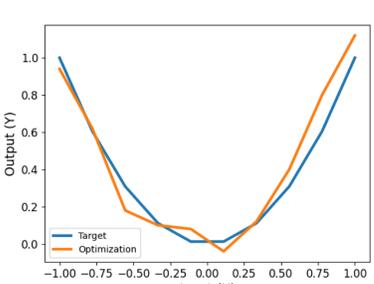 |

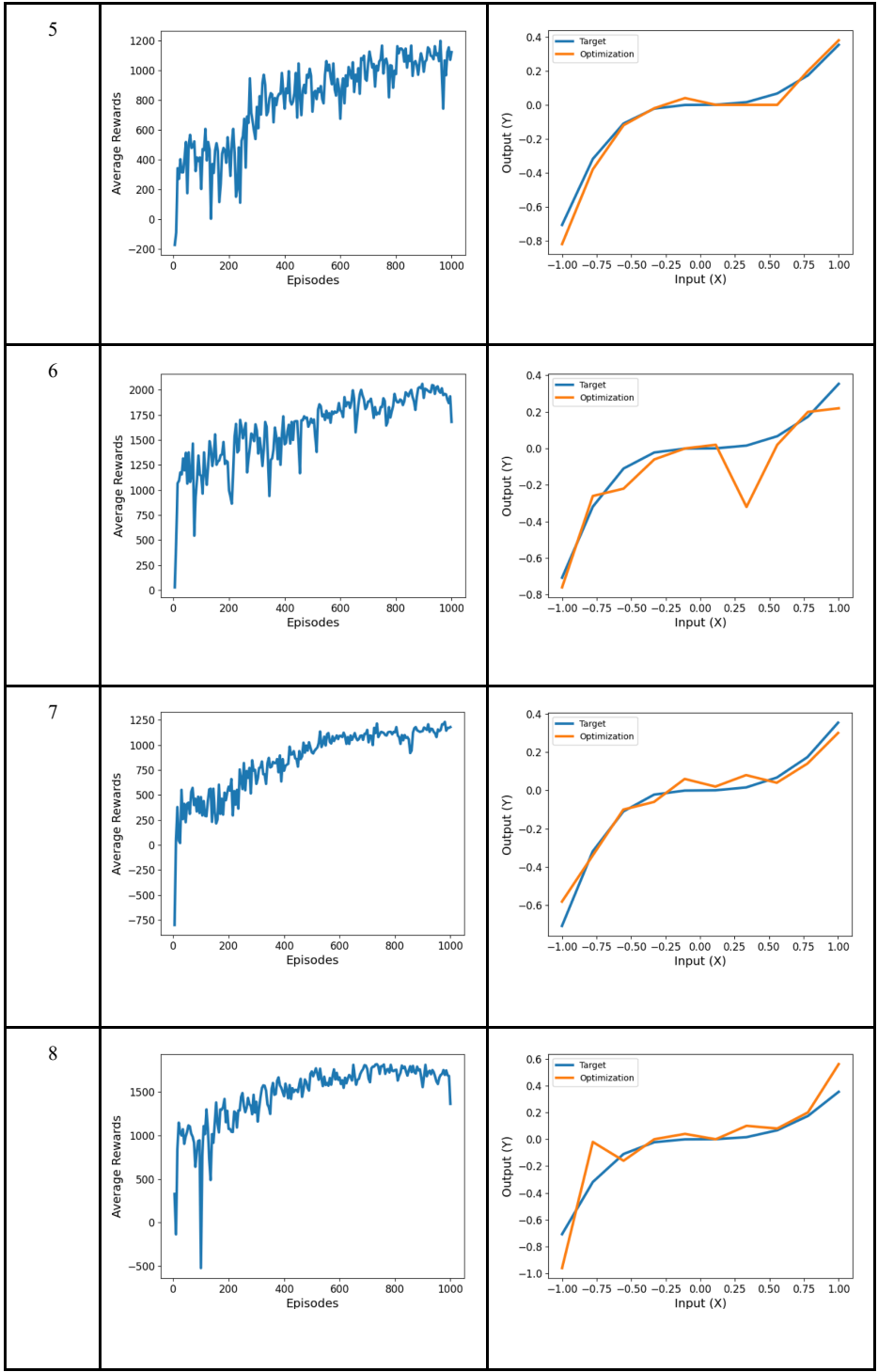

| 9 | 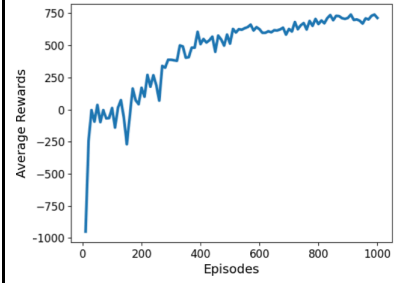 | 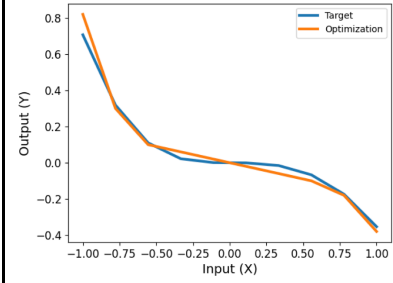 |